\documentclass[12pt]{article}

\usepackage[preprint, nonatbib]{neurips_2023}

\usepackage[utf8]{inputenc} 
\usepackage[T1]{fontenc}    
\usepackage{hyperref}       
\usepackage{url}            
\usepackage{booktabs}       
\usepackage{amsfonts, amsmath}       
\usepackage{nicefrac}       
\usepackage{microtype}      
\usepackage{xcolor}         

\usepackage{fancyhdr}

\usepackage{algorithm}
\usepackage{algpseudocode}

\hypersetup{colorlinks, linkcolor=blue, citecolor=blue, urlcolor=blue}

\usepackage{amsfonts,dsfont}
\usepackage{hyperref}
\usepackage{color}
\usepackage{amsmath, amssymb}
\usepackage{amsthm}
\usepackage{bm}
\usepackage{stmaryrd}
\usepackage{color}
\usepackage{comment}
\usepackage{geometry}
\allowdisplaybreaks

\usepackage{graphicx}
\usepackage{caption}
\usepackage{subcaption}

\usepackage{tabularx}

\usepackage{caption}

\newcommand{\todo}[1][\null]{\ensuremath{\clubsuit}}

\newcommand{\noprint}[1]{}

\theoremstyle{definition}
\theoremstyle{definition}
\theoremstyle{definition}
\theoremstyle{definition}
\theoremstyle{definition}
\theoremstyle{definition}
\theoremstyle{definition}
\theoremstyle{definition}
\theoremstyle{definition}

\usepackage{tikz}
\usetikzlibrary{positioning, arrows.meta, calc}

\title{Low-rank adaptive physics-informed HyperDeepONets for solving differential equations}

\author{%
  Etienne Zeudong \\
  African Institute for Mathematical Sciences\\
  Crystal Garden, P.O. Box 608, Limbe \\
  \texttt{etienne.zeudong@aims-cameroon.org} \\
  \And
  Elsa Cardoso-Bihlo \\
  Department of Mathematics and Statistics\\
  Memorial University of Newfoundland\\
  St. John’s, NL, A1C 5S7, Canada \\
  \texttt{ecardosobihl@mun.ca} \\
  \And
  Alex Bihlo \\
  Department of Mathematics and Statistics\\
  Memorial University of Newfoundland\\
  St. John’s, NL, A1C 5S7, Canada \\
  \texttt{abihlo@mun.ca} \\
}

\begin{document}

\maketitle

\vspace{0.25cm}

\begin{abstract}
HyperDeepONets were introduced in Lee, Cho and Hwang [ICLR, 2023] as an alternative architecture for operator learning, in which a hypernetwork generates the weights for the trunk net of a DeepONet. While this improves expressivity, it incurs high memory and computational costs due to the large number of output parameters required. In this work we introduce, in the physics-informed machine learning setting, a variation, PI-LoRA-HyperDeepONets, which leverage low-rank adaptation (LoRA) to reduce complexity by decomposing the hypernetwork's output layer weight matrix into two smaller low-rank matrices. This reduces the number of trainable parameters while introducing an extra regularization of the trunk networks' weights. Through extensive experiments on both ordinary and partial differential equations we show that PI-LoRA-HyperDeepONets achieve up to 70\% reduction in parameters and consistently outperform regular HyperDeepONets in terms of predictive accuracy and generalization.
\end{abstract}

\section{Introduction}

Solving differential equations with neural network based solution methods has become increasingly popular in the past several years~\cite{cuom22a}. There are two main approaches that have been put forth here, \textit{solution function approximation} methods and \textit{operator learning} methods. The first approach, originally proposed by Lagaris et al.~\cite{laga98a} rose to prominence in recent years thanks to the work of~\cite{rais19a}, dubbing this approach physics-informed neural networks. Here a neural network is trained to learn a single specific solution for a given set of parameters and initial/boundary conditions for the underlying system of differential equations. The second approach, introduced in~\cite{lu21a} rests on the universal approximation theorem for operators established by Chen and Chen~\cite{chen95a}, which guarantees that neural networks can approximate nonlinear operators under mild conditions. This approach learns the solution operator for the given system of differential equations, thus allowing generalization to sampling particular solutions corresponding to new parameters and different initial/boundary conditions without retraining of the network. While physics-informed neural networks are still more popular in many applications of science and engineering today, the operator learning approach is usually more powerful as it allows fast inference thus allowing speed-up of several orders of magnitude, even compared to standard numerical methods~\cite{li21a}.

Following the seminal work on data-driven deep operator networks~\cite{lu21a} and physics-informed deep operator networks~\cite{wang23a}, which use a neural network architecture based on a multiplicative combination of two sub-neural networks, referred to as branch and trunk networks, respectively, and directly inspired by the theoretical work of~\cite{chen95a}, many authors have proposed alternative approaches. These include the addition of further sub-networks~\cite{hado22a,vent23a} besides the branch and the trunk network, and different combination and weight generation strategies of the branch and trunk networks~\cite{lee23a,seid22a}. In particular, Lee et al.~\cite{lee23a} proposed \textit{HyperDeepONets}, which use the branch network to generate the weights of the trunk network, aiming to generalize the architectures proposed in~\cite{hado22a,seid22a,vent23a}. They have shown that HyperDeepONets can outperform these other DeepONet variations in a data-driven learning setting. However, an inherent problem of HyperDeepONets is their parameter complexity. As the branch network has to generate the weights of the trunk network, the output layer of the branch network has to have as many elements as there are parameters in the trunk network. As a consequence, this implies that the output layer of the branch network typically contains the vast majority of the trainable parameter of the branch network, and also means that scaling up the size of the trunk network requires increasing both the size and complexity of the branch network. To overcome these constraints, Lee et al.~\cite{lee23a} proposed a chunked version of HyperDeepOnets, which generates the target trunk weights in chunks rather than in one sweep. 

In this paper, we propose a new approach to mitigate the parameter inefficiency of HyperDeepONets by introducing low-rank adaptation (LoRA) into their architecture. Originally proposed for fine-tuning large language models~\cite{hu22a}, LoRA allows us to approximate the large output weight matrix of the hypernetwork as a product of two lower-dimensional matrices. This leads to a significant reduction of trainable parameters and an implicit regularization of the resulting model. As a further contribution, we illustrate this framework to the physics-informed setting, allowing us to learn solution operators directly from differential equations constraints without paired input--output data as is required for the data-driven setting considered in~\cite{lee23a}. Our numerical results show that LoRA-HyperDeepONets not only reduce memory and thus training costs, but also achieve lower errors than both standard DeepONets and full HyperDeepONets across a range of benchmarks of classical ordinary and partial differential equations.

The further outline for this paper is as follows. In Section~\ref{sec:RelatedWork} we provide a concise literature review on the relevant published works related to our proposed variation of a deep operator network. In Section~\ref{sec:HyperDeepONets} we describe the core idea of low-rank adaptive HyperDeepONets in the physics-informed learning setting. Numerical results for several important ordinary and partial differential equations comparing DeepOnets, HyperDeepONets and low-rank adaptive HyperDeepONets are presented in Section~\ref{sec:Results}. We finish this paper with the conclusions summarizing our findings in Section~\ref{sec:Conclusion}.

\section{Related work}\label{sec:RelatedWork}

Neural network based solution techniques for differential equations first emerged in~\cite{laga98a}, were popularized in~\cite{rais19a} and are since referred to as physics-informed neural networks. While the approach proposed in~\cite{laga98a,rais19a} is both conceptually simple and elegant, it suffers from several shortcomings, especially issues pertaining to computational cost~\cite{brec23b}, solution mode collapse for longer integration times~\cite{bihl22a,kris21a,wang23a,wang22b} and challenges regarding complex loss surfaces necessitating loss balancing strategies or more sophisticated optimization methods~\cite{bihl23a,mccl22a,wang22a}, as well as the inclusion of qualitative properties~\cite{aror23a,card23a} to obtain improved results.

While physics-informed neural networks learn a particular solution associated with a fixed initial--boundary value problem of a system of differential equations, and are thus classified as solution interpolation methods, it is also possible to learn the solution operator of the given system. For differential equations, the solution operator is a mapping from a space of initial--boundary conditions (defined over a function space) to an element of the solution space (also defined over a function space). Rather than learning a function interpolator, as is the case for physics-informed neural networks, operator learning aims to learn an operator interpolator as a surrogate for the true, often complex or analytically intractable, solution operator~\cite{chen95a,lee23a,lu21a}.

Operator learning holds great promise as a learned solution operator can be readily evaluated for new initial--boundary conditions without re-training a new neural network as is the case for physics-informed neural networks. This enables speed-ups by several orders of magnitude, making operator learning approaches computationally superior even to standard numerical methods. As such, they are actively researched for time-critical applications, such as weather forecasting or tsunami inundation modeling~\cite{bi23a,brec25a,koch24a}. 

Operator learning can be roughly classified into data-driven and physics-informed approaches. In data-driven approaches, a large dataset of paired input--output samples for the underlying system of differential equations is provided, and the solution operator is learned in a supervised learning setting. The main approaches for data-driven operator learning are Fourier neural operators~\cite{li21a} and DeepONets~\cite{lu21a}, see also~\cite{kova24a} for a review on operator learning. Physics-informed operator learning is often done in a semi-supervised or even fully unsupervised manner, depending on the availability of data and the specific problem setup. Rather than strictly requiring labeled input--output pairs to train the operator mapping, the solution operator is learned by enforcing a differential equation prior on the neural network's predictions~\cite{wang23a,wang21a}. This means the model is penalized if its output does not satisfy the given system of differential equations. While some approaches can indeed train an operator model using only the differential equation residual (unsupervised), many practical applications still benefit significantly from, or even require, a small amount of labeled data (semi-supervised) to constrain the solution and achieve higher accuracy, especially for complex or multi-modal solutions, as well as for solving inverse problems. See~\cite{li24a} for a review on physics-informed operator learning.

The success of operator learning approaches, both data-driven and physics-informed, has lead to a proliferation of architectures aiming at generalizing the standard DeepONet architecture proposed in~\cite{lu21a}. Some notable architectures are \textit{Shift-DeepONets}, proposed in~\cite{hado22a}, which add extra shift and scale networks to process the inputs of the branch net to be injected into the trunk network. The motivation for these extra sub-networks is to help DeepONets represent sharp features, by being able to learn such features in one location and then just shift and scale it to other locations; in other words, Shift-DeepONets aim to increase the expressivity of the basis for the trunk network, without a significant increase in overall parameter space. FlexDeepONets~\cite{vent23a} pursue a similar avenue as Shift-DeepONets by adding a pre-net conditioned on the inputs of the branch network to the trunk network. \textit{Nonlinear Manifold Decoder (NOMAD)}~\cite{seid22a} concatenate the outputs of the branch network to the trunk network aiming at making the DeepONet more expressive by allowing for a nonlinear reconstruction of the output function. HyperDeepONets~\cite{lee23a} replace the branch network itself with a hypernetwork, a neural network that generates the weights of another neural network~\cite{ha17a}, which generates the weights of the trunk network. It was argued in~\cite{lee23a} that rather than using the branch network to generate just a subset of all the weights for the trunk network, as is the case for Shift- and FlexDeepONets, all the weights of the trunk network should be generated by the branch network. A downside of this approach is that the branch network of HyperDeepONets has to generate as many outputs as there are weights in the trunk network and as such will have to involve large weight matrices. To overcome this limitation, the authors~\cite{lee23a} also proposed a chunked version of HyperDeepONets, which generates these weights in chunks rather than in one sweep. The downside of chunked HyperDeepONets is that they increase both the training times of the networks as well as memory consumption. They also introduce another hyper-parameter, the size of the chunk embedding.

\section{Low-rank adaptive physics-informed HyperDeepONets}\label{sec:HyperDeepONets}

In this section we present an overview of physics-informed DeepONets as proposed in~\cite{lu21a,wang23a}, summarize the architectural changes put forth in~\cite{lee23a} who introduced HyperDeepONets, and then present a low-rank adaptive version of these HyperDeepONets, the main contribution of our work.

\subsection{Physics-informed DeepONets}

We consider the problem of solving a system of differential equation defined over the spatio-temporal domain $\Omega\times[t_0, t_{\rm f}]$ as an initial--boundary value problem of the form
\begin{align}\label{eq:GeneralSystemOfDEs}
\begin{split}
&\Delta^l\big(t,x, u_{(n)}\big)=0,\quad l=1,\dots, L,\qquad t\in[t_0,t_{\rm f}],\ x\in\Omega,
\\[1ex]
&\mathsf I^{l_{\rm i}}\big(x, u_{(n_{\rm i})}\big|_{t=t_0}^{}\big)=0,\quad 
l_{\rm i}=1,\dots, L_{\rm i},\qquad x\in\Omega,
\\[1ex]
&\mathsf B_{\rm s}^{l_{\rm b}}\big(t, x, u_{(n_{\rm b})}\big)=0,\quad 
l_{\rm b}=1,\dots, L_{\rm b},\qquad t\in[t_0,t_{\rm f}],\ x\in\partial\Omega,
\end{split}
\end{align}
where $\Delta=(\Delta^1,\dots, \Delta^{L})$ is a system of differential equations of order $n$, $\mathsf I=(\mathsf I^1,\dots,\mathsf I^{L_{\rm i}})$ is an initial value operator 
and $\mathsf B_{\rm s}=(\mathsf B_{\rm s}^1,\dots,\mathsf B_{\rm s}^{L_{\rm b}})$ is a spatial boundary value operator. Here and in the following $x=(x_1,\dots, x_d)$ denotes the tuple of spatial independent variables, $\Omega\subset \mathbb{R}^d$, $\partial \Omega$ is the spatial boundary of $\Omega$, and $u=u(x) = (u^1,\dots, u^q)$ is the tuple of dependent variables. We use the notation from~\cite{olve86Ay} to denote by $u_{(n)}$ all the derivatives of the dependent variable $u$ with respect to the independent variables $(t,x)$ up to order $n$. In the following, we will only consider systems of evolution equations for which the initial value operator reduces to $\mathsf{I}=u|_{t=t_0}-u_0(x)$, and considering autonomous systems only we can set $t_0=0$ without restriction of generality. 

Formally, a solution operator for system~\eqref{eq:GeneralSystemOfDEs} is a mapping from the space of initial condition functions and boundary condition functions to the space of solution functions. Let $\mathcal {I}$ be the Banach space of initial condition functions and $\mathcal{B}$ be the Banach space of boundary condition functions. We denote the operator mapping from the combined space of these functions to the solution space $\mathcal U$ as
\[
\mathcal G\colon\quad  \mathcal{I}\times\mathcal{B} \to \mathcal U,\quad u=\mathsf{G}(u_0, g),
\]
where $u=u(t,x)=\mathsf{G}(u_0, g)(t,x)$ is the function satisfying system~\eqref{eq:GeneralSystemOfDEs} with the initial condition determined by $u_0(x)$ and the boundary condition determined by $g(t,x)$ as formally expressed by the boundary value operator $\mathsf{B}_{\rm s}$. To simplify the notation, we will omit the explicit dependency of the solution operator $\mathsf{G}$ on $g$ in the following. This is justified, as in all the examples that follow we design our neural network solution operators over the class of functions that identically satisfy the boundary conditions. 

Physics-informed DeepONets then aim to approximate the solution operator of system~\eqref{eq:GeneralSystemOfDEs} using a deep neural network. We denote the approximation to $\mathsf{G}$ using a neural network with weights $\theta$ as $\mathsf{G}^\theta$. A standard DeepONet as proposed in~\cite{lu21a} consists of two sub-networks, a \textit{branch network} for the input function and a \textit{trunk network} for the domain coordinates. Specifically, the initial condition $u_0(x)$ is sampled using a discrete representation at a fixed number of sensor locations, denoted by $n_{\rm sensors}$ over the domain $\Omega$. This discrete representation $\hat u_0=(u_0(x_1),\dots u_0(x_{\rm sensor}))$ becomes the input to the branch network. The branch network itself is typically a fully connected neural network, although other types of networks such as convolutional or graph neural network can be considered as well, especially for higher-dimensional problems or equations formulated on general unstructured domains. The output of the branch network is a vector of coefficients, $B=(B_1,\dots B_p)$ which is the latent representation of the initial condition function. Here $p$ is a hyper-parameter, the dimensionality of the latent embedding. The trunk network processes the independent variables of the solution domain, thus accepting the spatio-temporal coordinates $(t,x)$ where one wants to evaluate the solution as an input. The output of the trunk network is the vector of basis functions $T(t,x)=(T_1(t,x),\dots, T_p(t,x))$, with $p$ matching the output dimension of the branch network. A standard DeepONet then approximates the solution operator by taking the dot product of the outputs from the branch and the trunk networks:
\[
u(t,x) \approx u^{\theta}(t,x) = \mathsf{G}^\theta(\hat u_0)(t,x) = \sum_{k=1}^p B_k T_k(t,x) = B\cdot T(t,x). 
\]

DeepOnets can be trained both in a data-driven~\cite{lu21a} and physics-informed manner~\cite{wang21a}. We focus on the physics-informed case here only. Physics-informed DeepONets penalize the network if the output $u^\theta(t,x)$ violates the system of differential equations~\eqref{eq:GeneralSystemOfDEs}. To accomplish this, a physics-informed loss function is designed as
\begin{gather}\label{eq:compositeLossFunctionIBVP}
\mathcal L(\theta) = 
 \mathcal L_\Delta(\theta) 
+\lambda_{\mathsf I}\mathcal L_{\mathsf I}(\theta)
+\lambda_{\mathsf B}\mathcal L_{\mathsf B}(\theta),
\end{gather}
with
\begin{align*}
\mathcal L_\Delta(\theta) ={}&\frac1{N_{\rm u}N_\Delta}\sum_{i=1}^{N_\Delta}\sum_{j=1}^{N_{\rm u}}\sum_{l=1}^L
\big|\Delta^l\big(t^i_\Delta,x^i_\Delta,u^{j,\theta}_{(n)}(t^i_\Delta,x^i_\Delta)\big)\big|^2,
\quad (t^i_\Delta,x^i_\Delta)\in[t_0,t_{\rm f}]\times\Omega,\\[1.5ex]
\mathcal L_{\mathsf I}(\theta) ={}& \frac1{N_{\rm u}N_{\rm i}}\sum_{i=1}^{N_{\rm i}}\sum_{j=1}^{N_{\rm u}}\sum_{l_{\rm i}=1}^{L_{\rm i}}
\big|\mathsf I^{l_{\rm i}}\big(x^i_{\rm i},u_{(n_{\rm i})}^{j,\theta}(t_0,x^i_{\rm i})\big)\big|^2,
\quad x^i_{\rm i}\in\Omega,
\\[1.5ex]
\mathcal L_{\mathsf B}(\theta) ={}& \frac1{N_{\rm u}N_{\rm b}}\sum_{i=1}^{N_{\rm b}}\sum_{j=1}^{N_{\rm u}}\sum_{l_{\rm b}=1}^{L_{\rm b}}
\big|\mathsf B_{\rm s}^{l_{\rm b}}\big(t^i_{\rm b},x^i_{\rm b},u_{(n_{\rm b})}^{j,\theta}(t^i_{\rm b},x^i_{\rm b})\big)\big|^2,
\quad (t^i_{\rm b},x^i_{\rm b})\in[t_0,t_{\rm f}]\times\partial\Omega,
\end{align*}
being the system of differential equations residual loss, the initial value loss and the boundary value loss, and $\lambda_{\mathsf I}$ and $\lambda_{\mathsf B}$ are positive weights, which may change over the course of training the physics-informed DeepONet~\cite{mccl22a}. These mean squared errors are computed at the sets of collocation points 
$\{t_\Delta^i, x_\Delta^i\}_{i=1}^{N_\Delta}\subset[t_0,t_{\rm f}]\times \Omega$, $\{x_{\rm i}^i\}_{i=1}^{N_{\rm i}}\subset \Omega$ and 
$\{x_{\rm b}^i\}_{i=1}^{N_{\rm b}}\subset\partial\Omega$
for the system of differential equations, the initial conditions and the boundary conditions, respectively. We denote $u^{j,\theta}=\mathsf{G}^\theta(\hat u^j_0) $, and the set of discretized initial values $\{\hat u^j_0\}_{j=1}^{N_{\rm u}}$ are sampled from a suitably expressive class of basis functions, such as truncated Fourier series, Chebyshev polynomials or Gaussian random fields, for which the solution operator $\mathsf{G}(u_0)$ should be learned~\cite{brec23a,lu21a,wang21a}.

The physics-informed loss function~\eqref{eq:compositeLossFunctionIBVP} is typically being minimized using first-order (stochastic) optimization, although higher-order methods such as L-BFGS have been used as well in the literature.

As indicated above, it is generally advisable to enforce the initial and boundary conditions as a hard constraint, see e.g.~\cite{laga98a} and~\cite{brec23a}, which would allow to simplify the composite loss function~\eqref{eq:compositeLossFunctionIBVP}. In the following partial differential examples we consider periodic boundary conditions, which can be hard constrained in the network by including a coordinate-transform layer as first layer after the input layer to the trunk network that transforms in terms of sine and cosine functions~\cite{bihl22a}. This guarantees that $\mathcal L_{\mathsf B}(\theta)$ is automatically zero and hence can be removed from the composite loss function~\eqref{eq:compositeLossFunctionIBVP}. The initial condition can be hard constrained in the solution operator network as well using, e.g., the ansatz
\[
\mathsf G^\theta(u_0)(t,x) = u_0(x) + t\mathsf G_{\rm train}^\theta(u_0)(t,x),
\]
where $\mathsf{G}^\theta_{\rm train}$ is the trainable part of the physics-informed DeepONet $\mathsf G^\theta(u_0)(t,x)$. This ansatz guarantees that the initial value loss $\mathcal L_{\mathsf I}(\theta)$ is zero as well, and the physics-informed loss function~\eqref{eq:compositeLossFunctionIBVP} reduces to the differential equations residual loss only. More complicated hard constraints can be used as well, and the reader is referred to~\cite{brec23a} for further details.

\subsection{HyperDeepONets}

HyperDeepONets were introduced in~\cite{lee23a} who presented them in a data-driven setting. That is, a dataset of input--output pairs was constructed, and the HyperDeepONet was then trained to learn the implicit operator representation for these input--output pairs. A HyperDeepONet replaces the the branch network in the standard DeepONet with a hypernetwork~\cite{ha17a} that generates the required weights of the trunk network. That is, the inputs to the hypernetwork are the initial value functions sampled at the $n_{\rm sensor}$ sensor points, and the output generates the $n_{\rm trunk}\gg 1$ trainable parameters that make up all the weights and biases of all layers in the trunk network. As in the standard DeepONet, the hypernetwork branch net of the HyperDeepONet also uses a fully connected neural network. 

The implication of this architectural choice is that the output layer of the branch net hypernetwork is typically large, as the trunk network of DeepONets typically requires several thousand or more parameters. We denote the output layer of the branch net hypernetwork as $y^{\rm branch}_{\rm out}=W^{\rm branch}_{\rm out}x+b^{\rm branch}_{\rm out}$, where $x$ is the input to the last hidden layer of size $n_{\rm hidden}$ of the hypernetwork. That is, the weight matrix $W^{\rm branch}_{\rm out}$ and the bias vector $b^{\rm branch}_{\rm out}$ are elements of $\mathbb{R}^{n_{\rm trunk}\times n_{\rm hidden}}$ and $\mathbb{R}^{n_{\rm trunk}}$, respectively. In general, most of the trainable parameters of the branch net of the HyperDeepONet will be in $W^{\rm branch}_{\rm out}$, making this network extremely unbalanced, which increases the chance of overfitting.

\subsection{Low-rank adaptive HyperDeepONets}

\begin{figure}[ht!]
    \centering
    \begin{minipage}{0.48\textwidth} 
        \centering
        \includegraphics[width=\textwidth]{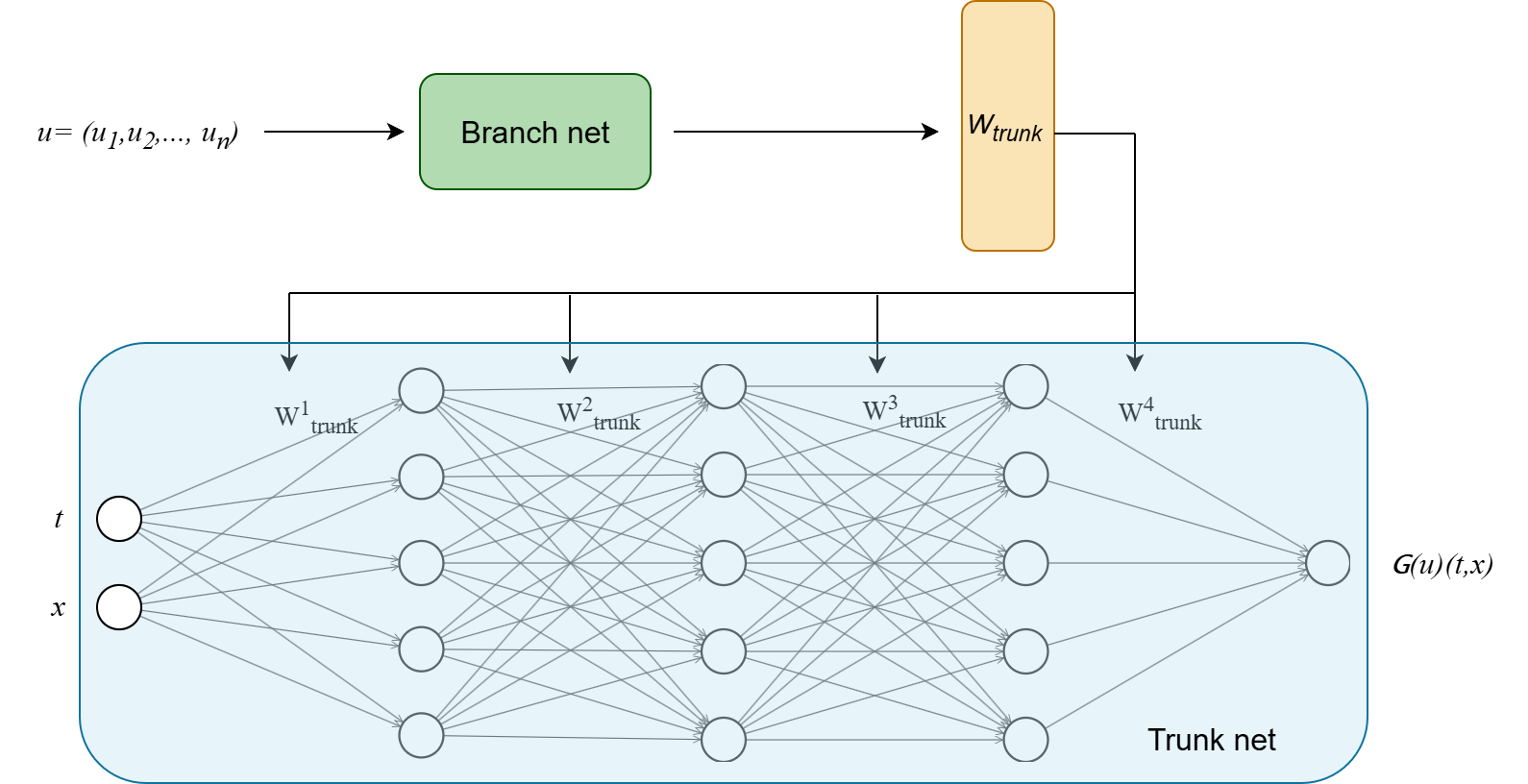} 
    \end{minipage}%
    \hfill%
    \begin{minipage}{0.48\textwidth}
        \centering
        \includegraphics[width=\textwidth]{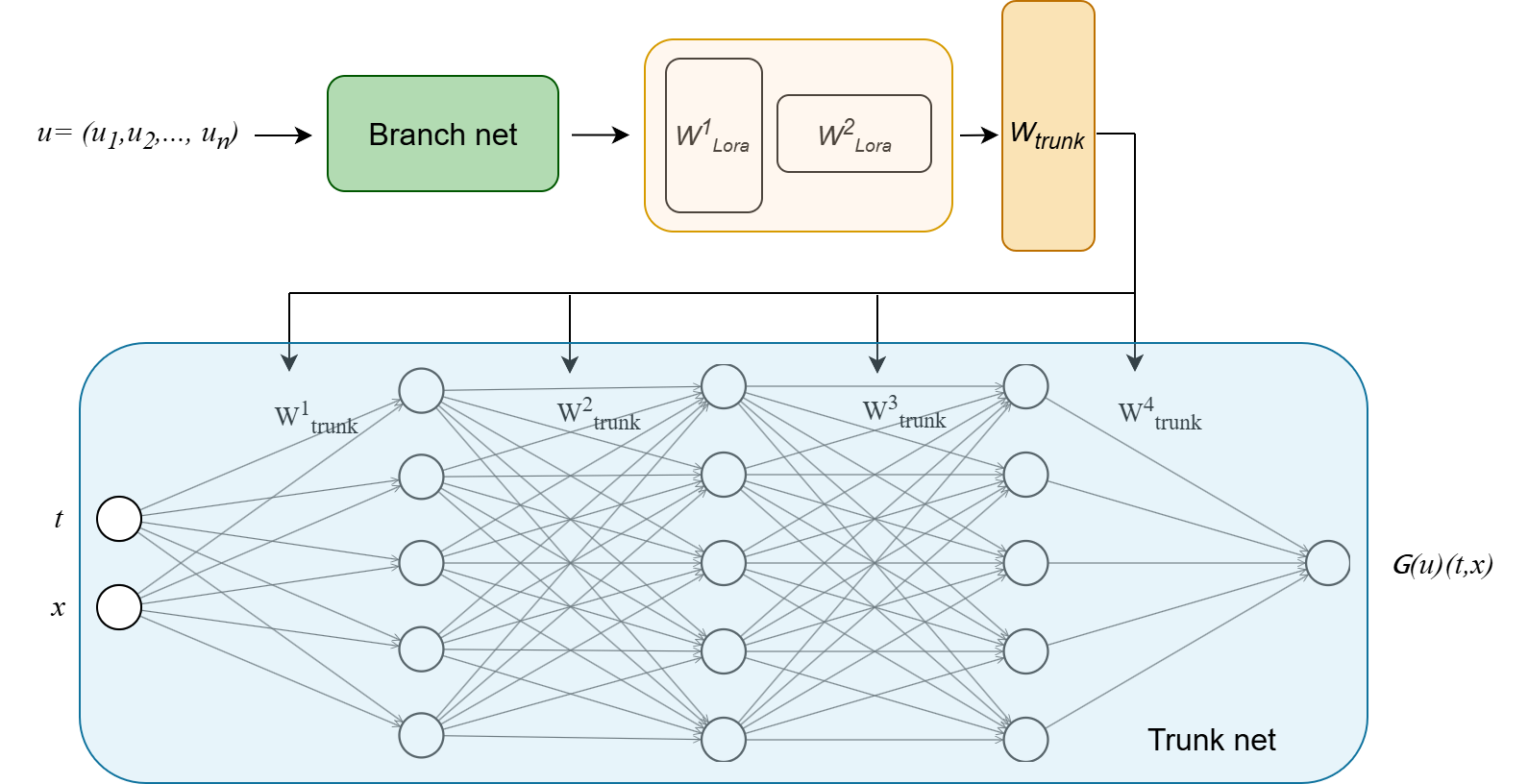}
    \end{minipage}
    \caption{Illustration of the architectures of a full HyperDeepONet \textit{(left)} and of the proposed LoRA-HyperDeepONet \textit{(right)} for the example of an operator learning problem in the two independent variables $(t,x)$ and one dependent variable $u$. The weight matrix connecting the last hidden layer with the output layer of the branch net's hyper-network is a full matrix of size $n_{\rm trunk}\times n_{\rm hidden}$, which in the LoRA-version is decomposed as two low-rank matrices of sizes $n_{\rm trunk}\times r$ and $n_r\times n_{\rm hidden}$, respectively, with $r<n_{\rm hidden}\ll n_{\rm trunk}$.}
    \label{fig:HyperLoRa}
\end{figure}

The extension of HyperDeepONets to low-rank adaptive HyperDeepONets is straightforward, and follows the main idea behind low-rank adaptation as used to fine-tune large language models~\cite{hu22a}. Rather than directly computing $W^{\rm branch}_{\rm out}$, we set
\begin{equation}\label{eq:LoRA}
    W^{\rm branch}_{\rm out, Lora} = W^1_{\rm Lora}W^2_{\rm Lora},
\end{equation}
where $W^1_{\rm Lora} \in\mathbb{R}^{n_{\rm trunk}\times r}$ and $W^2_{\rm Lora} \in\mathbb{R}^{r\times n_{\rm hidden}}$ are two matrices of rank $\min(r, n_{\rm trunk})$ and $\min(r,n_{\rm hidden})$. Since in general $n_{\rm hidden}\ll n_{\rm trunk}$ we choose $r<n_{\rm hidden}$ so that both matrices $W^1_{\rm Lora}$, and $W^2_{\rm Lora}$ have rank at most $r$. In this case, the low-rank adaptive version of the hypernetwork's branch net output layer weight matrix $W^{\rm branch}_{\rm out, Lora}$ reduces the number of trainable parameters of $W^{\rm branch}_{\rm out}$ from $n_{\rm trunk}\times n_{\rm hidden}$ to $r\times(n_{\rm trunk}+n_{\rm hidden})$. This also implies that the weights of all the layers of the trunk network generated by the LoRA-hypernetwork branch net are not independent of one another, which implies an extra coupling that is not present in the original version of the HyperDeepONet, and which implicitly adds a regularization to the network. We illustrate in the following section with several examples that this regularization can actually lead to better performance of the LoRA-HyperDeepONets compared to the full HyperDeepONets, with the added benefit of in general requiring much fewer trainable parameters. A graphical illustration of full HyperDeepONets and LoRA-HyperDeepONets is given in Figure~\ref{fig:HyperLoRa}.

\section{Numerical results}\label{sec:Results}

In this section we present several numerical examples, highlighting the performance of low-rank adaptive HyperDeepONets. We consider both ordinary and partial differential equations. Where possible, the architectures used are inspired by those in similar examples shown in~\cite{lee23a}. Note that~\cite{lee23a} considers only the data-driven case, whereas here we consider the physics-informed case. That is, no data is being used to train the DeepONets, besides initial conditions sampled from suitable classes of initial value problems, which will be detailed in each of the following examples.

All DeepONets use fully connected feedforward neural networks for both the branch net and trunk net. The hyperbolic activation functions is used for all hidden layers throughout all networks, and the last output layer uses a linear activation function. The partial differential equations examples all use $n_{\rm sensor}=128$ regularly spaced sensor points to sample the initial conditions. For each example we generate $N^{\rm train}_{\rm u}=10,000$ training samples with the associated classes of initial conditions for which the solution operator is being learned being described in the respective example subsection. We also use $N_{\Delta}=10,000$ differential equations collocation points, and, if applicable, $N_{\rm i}=10,000$ initial value collocation points. The neural networks are then tested on $N^{\rm test}_{\rm u}=100$ samples, and the relative $l_2$-errors with associated standard deviations are reported over the test dataset. To obtain robust statistics, each DeepONet variation is trained a total of 10 times with randomly initialized weights, with all $l_2$-errors reported below being averaged across all 10 models. 

A key challenge in comparing the performance of standard DeepONets and HyperDeepONets is their fundamentally different architectures, which makes a direct one-to-one comparison of network structures difficult. To create a fair comparison, our guiding principle is to ensure that both models have roughly the same number of total trainable parameters, with the primary distinction being how these parameters are allocated between the branch and trunk networks. This still allows for considerable flexibility in the specific configurations. In the experiments that follow, we explore several strategies to achieve this parameter parity. These approaches include using identical trunk network architectures while adjusting the branch net sizes, adopting architectures inspired by those in Lee et al.~\cite{lee23a} for similar problems, and even swapping the parameter distribution between the branch and trunk networks to analyze the impact of these structural choices on performance. To ensure a fair initialization, the weights for both network types are set so that their initial loss values are within an order of magnitude of each other before training commences. For the HyperDeepONet and its LoRA variations, the weight matrices $W^1_{\rm Lora}$ and $W^2_{\rm Lora}$ are initialized so that the elements of $W^{\rm branch}_{\rm out, Lora}$ and $W^{\rm branch}_{\rm out}$ have the same variance. Where possible, which is for the harmonic oscillator and the linear advection equation examples, we use the exact solution as a reference solution; for all other examples numerical reference solutions are obtained using a spectral discretization in space and a Runge--Kutta time-stepper employing the method of lines to obtain a high fidelity numerical reference solution~\cite{durr10a}.

The Adam optimizer with a fixed learning rate of $\eta=10^{-3}$ is used for all experiments, using a batch size of $N^{\rm train}_{\rm u}$. All DeepONets were implemented in \texttt{JAX}~\cite{brad18a} and trained on a single NVIDIA RTX 4090 GPU. The codes for reproducing the results can be found on \texttt{GitHub} upon publication of the article.\footnote{\url{https://github.com/abihlo/PI-HyperDeepONets}}

\subsection{Harmonic oscillator}

We begin by investigating the simple harmonic oscillator, in the Hamiltonian representation
\begin{equation}\label{eq:HarmonicOscillator}
    \frac{\mathrm{d}q}{\mathrm{d}t} = -\frac{p}{m},\qquad \frac{\mathrm{d}p}{\mathrm{d}t} = kq,
\end{equation}
where we set the mass $m=1$ and the spring constant $k=1$. The initial conditions for $q_0$ and $p_0$ were sampled from random uniform distributions over the interval $[-1,1]$. All DeepONets are trained over the temporal interval $[0,t_{\rm f}]$ with $t_{\rm f} = \pi$, which is half a period of the solution for the harmonic oscillator for the chosen values of $k$ and $m$.
 
The standard DeepONet uses a large branch network of 4 layers with 80 units per layer, and a small trunk network of 2 layers with 10 units per layer. The HyperDeepONet in turn uses a small branch network with 2 layers and 10 units per layer and a large trunk network of 4 layers and 20 units per layer. As listed in Table~\ref{tab:HarmonicOscillator} this choice leads to roughly the same number of trainable parameters between the standard DeepONet and the full HyperDeepONet. We enforce all initial conditions as hard constraints. 

One of the main features of DeepONets is that they allow for time-stepping for autonomous systems of equations by using the final solution at time $t_{\rm f}$ over one step as the initial condition for the next time step, as long as this final solution still falls within the training range of the DeepONet~\cite{wang23a}. To assess the generalization capabilities of the DeepONets for the harmonic oscillator we perform both a one-step prediction over the same time interval over which the networks were trained, and an iterative 10-step prediction to investigate the handling of error accumulations of all networks.

\begin{table}[!ht]
\centering
\caption{Numerical results for the harmonic oscillator~\eqref{eq:HarmonicOscillator} evaluated over the testing dataset.}\label{tab:HarmonicOscillator}
\begin{tabular}{llrrll}
\toprule
Model & LoRA Rank & Params & One-step $l_2$-error & 10-step $l_2$-error \\
\midrule
DeepONet & N/A & 15060 & $0.0099 \pm 0.0062$ & $0.076 \pm 0.048$ \\
HyperDeepONet & N/A & 14902 & $0.0119 \pm 0.0203$ & $0.075 \pm 0.089$ \\
LoRA-HyperDeepONet & 2 & 4186 & $0.0070 \pm 0.0061$ & $0.046 \pm 0.036$ \\
LoRA-HyperDeepONet & 4 & 6890 & $\textbf{0.0067} \pm \textbf{0.0060}$ & $\textbf{0.045} \pm \textbf{0.036}$ \\
LoRA-HyperDeepONet & 6 & 9594 & $0.0079 \pm 0.0069$ & $0.055 \pm 0.042$ \\
\bottomrule
\end{tabular}
\end{table}

Table~\ref{tab:HarmonicOscillator} contains the numerical results of our parameter study. It shows that for the simple example of the harmonic oscillator, the HyperDeepONet is not able to outperform the standard DeepONet for a one-step prediction and gives roughly the same error for the 10-step prediction case. This is not surprising, as the HyperDeepONet proposed in~\cite{lee23a} was specifically designed for problems that give rise to complex target solutions, which the simple harmonic oscillator does not exhibit. Interestingly though, the LoRA-HyperDeepOnet with ranks as low as $r=2$ can outperform both the full HyperDeepONet and the standard DeepONet, both for one-step and for 10-step predictions, using less than 30\% of the parameters of these other networks. Indeed, all the LoRA-HyperDeepONet can outperform these networks, with the best result obtained for rank $r=4$ here, at a fraction of the number of trainable parameters needed. 

\begin{figure}[!ht]
    \centering
    \includegraphics[width=0.9\linewidth]{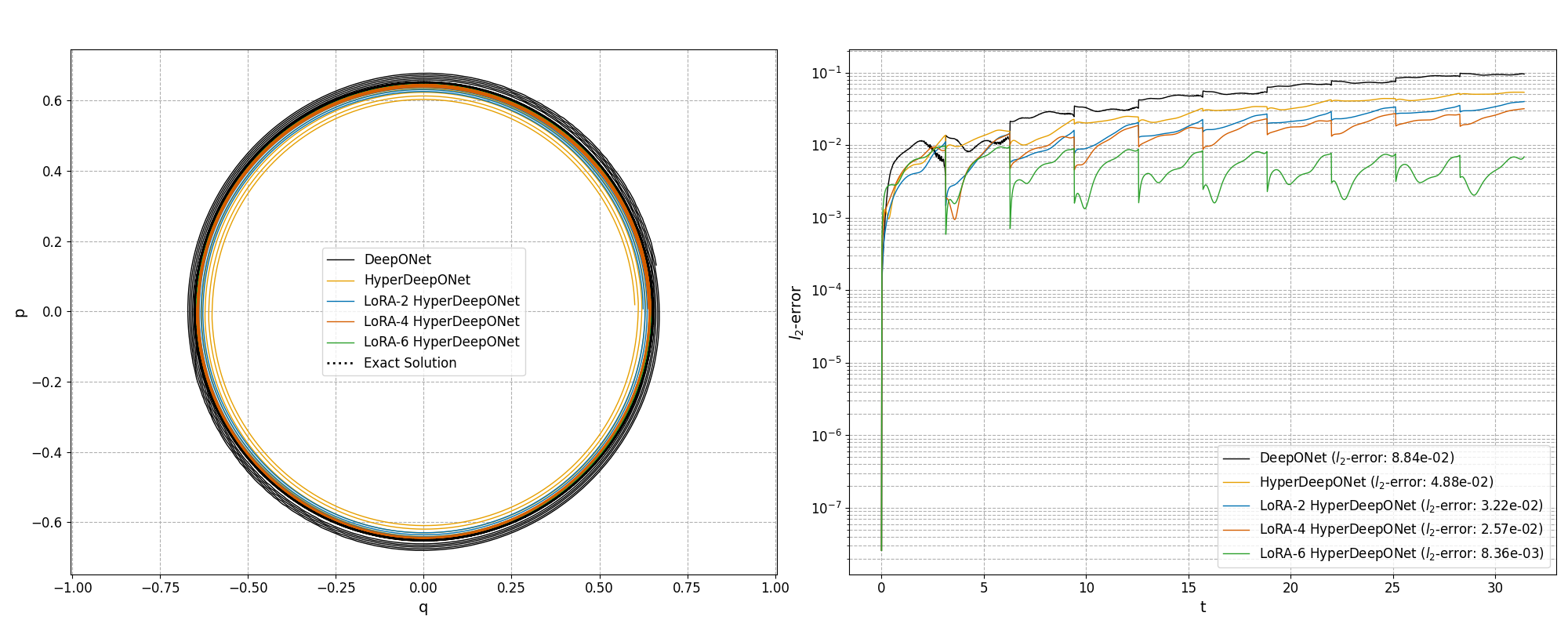}
    \caption{A case study for the harmonic oscillator~\eqref{eq:HarmonicOscillator} for all the trained DeepONets, evaluated over 10 steps starting from a random initial condition $(q_0,p_0)$ drawn from $[-1,1]^2$. The exact solution completes 5 full periods over this time interval. \textit{Left:} Phase plot. \textit{Right:} Associated $l_2$-error time series.}
    \label{fig:HarmonicOscillator}
\end{figure}

Figure~\ref{fig:HarmonicOscillator} shows a case study comparing the trained DeepONets against the exact solution for a randomly sampled initial condition $(q_0,p_0)$ drawn from the interval $[-1,1]^2$. We note that the standard DeepONet and the full HyperDeepONet exhibit a rather strong spurious drift and thus do not as accurately capture the long-time behaviour of the underlying periodic solution as the LoRA-HyperDeepONet variations do.

\subsection{Rigid body equations}

We consider the Euler equations for the free rigid body. Using angular momentum representation for this system, the governing equations are
\begin{align}\label{eq:RigidBody}
\begin{split}
   &\frac{\mathrm{d}\omega_1}{\mathrm{d}t} = \frac{I_2-I_3}{I_2I_3}\omega_2\omega_3,\\
   &\frac{\mathrm{d}\omega_2}{\mathrm{d}t} = \frac{I_3-I_1}{I_1I_3}\omega_1\omega_3,\\
   &\frac{\mathrm{d}\omega_3}{\mathrm{d}t} = \frac{I_1-I_2}{I_1I_2}\omega_1\omega_2,
\end{split}
\end{align}
where $\omega=(\omega_1,\omega_2,\omega_3)$ is the angular momentum in the rigid body frame and $I_1$, $I_2$ and $I_3$ are the principal moments of inertia~\cite{hair06Ay,holm09a}. We set $I_1=1.0$, $I_2=2.0$ and $I_3=3.0$. The initial conditions for $\omega_1$, $\omega_2$ and $\omega_3$ are sampled from a uniform random distribution over the interval $[-1,1]$. All networks are trained to approximate the solution over the temporal interval $[0,t_{\rm f}]$, with $t_{\rm f}=1.0$.

The standard DeepONet has a trunk network of 4 layers with 20 units per layer, and the branch network has 4 layers and 102 units per layer. The HyperDeepONet uses the same trunk network architecture as the standard DeepONet and a branch network with 4 layers and 20 units. This ensures again that both standard and HyperDeepONet have roughly the same number of trainable parameters. As in the previous case, the initial conditions are enforced as a hard constraint.

\begin{table}[!ht]
\centering
\caption{Numerical results for the rigid body~\eqref{eq:RigidBody} evaluated over the testing dataset.}\label{tab:RigidBody}
\begin{tabular}{llrll}
\toprule
Model & LoRA Rank & Params & One-step $l_2$-error & 10-step $l_2$-error \\
\midrule
DeepONet & N/A & 29740 & $0.0052 \pm 0.0037$ & $0.0469 \pm 0.0272$ \\
HyperDeepONet & N/A & 29963 & $0.0012 \pm 0.0006$ & $0.0110 \pm 0.0035$ \\
LoRA-HyperDeepONet & 4 & 8235 & $\textbf{0.0010} \pm \textbf{0.0003}$ & $\textbf{0.0107} \pm \textbf{0.0034}$ \\
LoRA-HyperDeepONet & 8 & 13767 & $0.0012 \pm 0.0006$ & $0.0112 \pm 0.0036$ \\
LoRA-HyperDeepONet & 16 & 24831 & $0.0012 \pm 0.0006$ & $0.0120 \pm 0.0065$ \\
\bottomrule
\end{tabular}
\end{table}

Table~\ref{tab:RigidBody} collects the evaluation results for this test case. In contrast to the harmonic oscillator, here the full HyperDeepONet can outperform the standard DeepONet. However, as in the previous example also for the rigid body the LoRA-HyperDeepONet with a rank of just 4 can outperform the full HyperDeepONet, again using less than 30\% of the parameters of both the full HyperDeepONet and the standard DeepONet. This example corroborates the findings of~\cite{lee23a} that HyperDeepONets outperform standard DeepONets, while we in addition show that using low-rank adaptation within HyperDeepONets can further improve the results that can be achieved using the framework of hyper-networks.

\begin{figure}[!ht]
    \centering
    \includegraphics[width=0.9\linewidth]{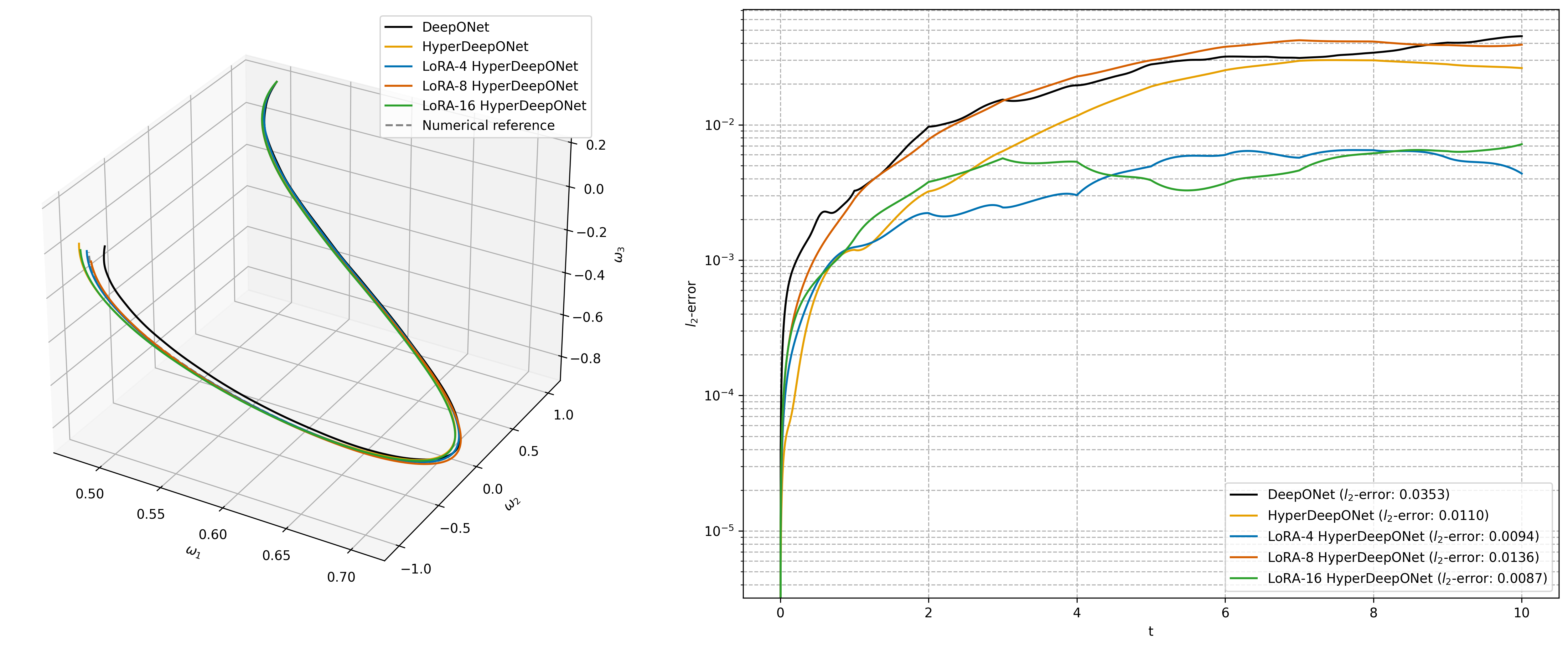}
    \caption{A case study for the rigid body equations~\eqref{eq:RigidBody} for all the trained DeepONets, evaluated over 10 steps starting from a random initial conditions $(\omega_1,\omega_2,\omega_3)_0$ drawn from $[-1,1]^3$. \textit{Left:} Phase plot. \textit{Right:} Associated $l_2$-error time series.}
    \label{fig:RigidBody}
\end{figure}

Figure~\ref{fig:RigidBody} presents a case study showing the time series of a 10-step iterative prediction starting from a randomly sampled initial condition $(\omega_1,\omega_2,\omega_3)_0$ with each component drawn from the range $[-1,1]$. All of the HyperNetworks, both full and low-rank adaptive, outperform the standard DeepONet for this test case. The lowest overall $l_2$-error for this particular example is achieved by the rank 16 approximation, closely followed by the rank 4 approximation, the full HyperDeepONet and the rank 8 HyperDeepONet, with the standard DeepONet performing worst among all models. The error per trainable parameter count is again much lower for the LoRA-HyperDeepONets than for the full HyperDeepONet and the standard DeepONet.

\subsection{Linear advection equation}

After having considered two examples of ordinary differential equations, we now move on to partial differential equations. As a first example, here we consider the linear advection equation
\begin{equation}\label{eq:LinearAdvection}
u_t + cu_x = 0,
\end{equation}
with constant advection velocity $c=1$, and using periodic boundary conditions on the domain $[-\pi,\pi]$, integrated over $[0,t_{\rm f}]$ with $t_{\rm f}=1$. The initial conditions are obtained from a truncated Fourier series expansion, and we use periodic boundary conditions that are enforced using hard constraints as proposed in~\cite{bihl22a}. The initial conditions are enforced as hard constraint as well, as was proposed in~\cite{brec23a}.

The architectures of the networks are as follows. The standard DeepONet uses a total of 4 layers for the trunk network with 256 units per layer, and a branch network with 2 layers and 256 units per layer. The HyperDeepONet uses a trunk network with 4 layers and 33 units per layer, and a branch network with 5 layers and 70 units per layer. This architecture was chosen to closely follow the architecture used in~\cite{lee23a} where they used the same equation in the purely data-driven case.

\begin{table}[!ht]
\centering
\caption{Numerical results for the linear advection equation~\eqref{eq:LinearAdvection} over the testing dataset.}\label{tab:LinearAdvection}
\begin{tabular}{llrl}
\toprule
Model & LoRA Rank & Params & $l_2$-error \\
\midrule
DeepONet & N/A & 297216 & $0.0800 \pm 0.0348$ \\
HyperDeepONet & N/A & 279682 & $0.0483 \pm 0.0054$ \\
LoRA-HyperDeepONet & 4 & 46850 & $0.0362 \pm 0.0097$ \\
LoRA-HyperDeepONet & 8 & 61258 & $\textbf{0.0284} \pm \textbf{0.0051}$ \\
LoRA-HyperDeepONet & 16 & 90074 & $0.0317 \pm 0.0086$ \\
\bottomrule
\end{tabular}
\end{table}

Table~\ref{tab:LinearAdvection} contains the error metrics for the trained DeepONets evaluated over the testing dataset. Similar as was found in~\cite{lee23a} for the data-driven case, the HyperDeepONet also outperforms the standard DeepOnet in the physics-informed setting. What is more, as in the previous examples for systems of ordinary differential equations, the low-rank adaptive versions of HyperDeepONets further improve upon the full HyperDeepONet also for the case of the linear advection equation. We found that the rank 8 approximation gives the overall lowest errors while only using roughly 20\% of the number of parameters of both the full HyperDeepONet and the standard DeepONet. For the example of the linear advection equation, all of the low-rank adaptive HyperDeepONets outperform the full HyperDeepONet.

\begin{figure}[!ht]
    \centering
    \includegraphics[width=\linewidth]{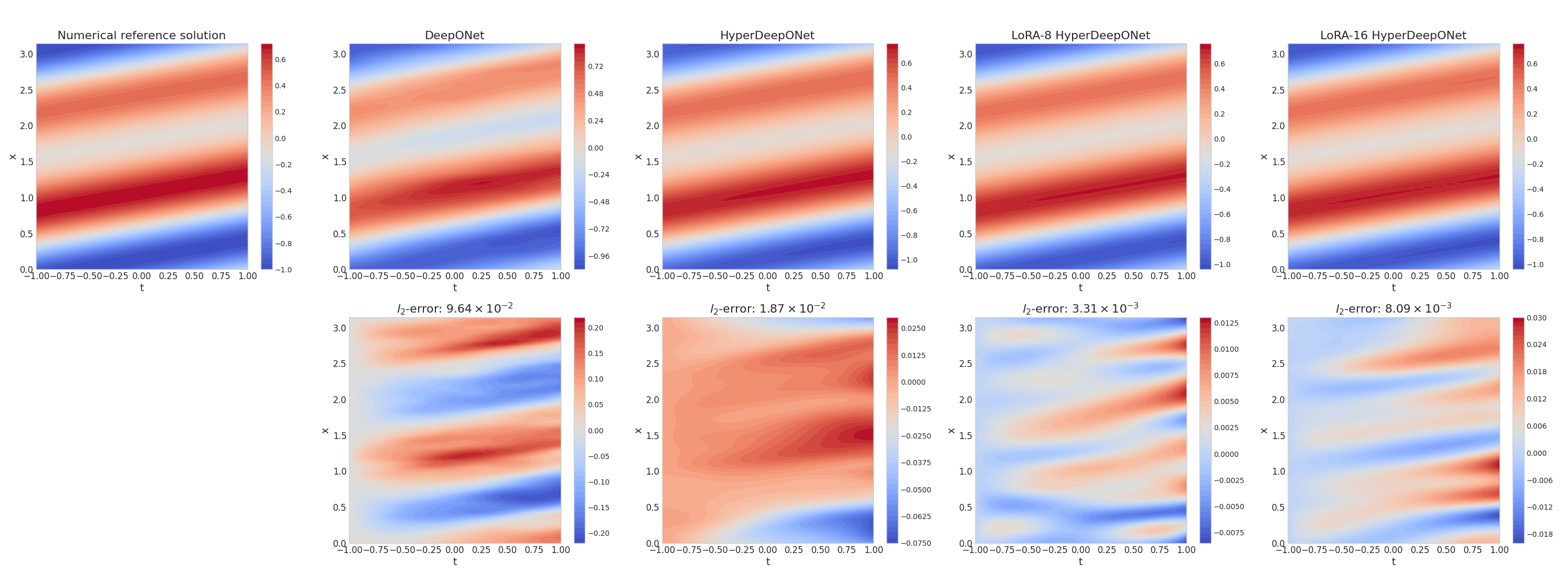}
    \caption{A case study for the linear advection equation~\eqref{eq:LinearAdvection} for all the trained DeepONets evaluated over a single time-step starting from a randomly sampled initial condition $u_0(x)$ obtained from a truncated Fourier series expansion. \textit{Top row:} Numerical reference solution and DeepONet solutions. \textit{Bottom row:} Point-wise differences between the reference solution and the DeepONet solutions.}
    \label{fig:LinearAdvection}
\end{figure}

A case study corresponding to a randomly sampled initial condition $u_0(x)$ is presented in Figure~\ref{fig:LinearAdvection}. As implied by the aggregated test results presented in Table~\ref{tab:LinearAdvection}, the HyperDeepONets improve upon the numerical results achievable with the standard DeepOnet for this particular test case setup. The standard DeepONet struggles to correctly advect the initial wave pattern and spuriously dissipates the solution over the course of the time interval. The HyperDeepONets do not suffer from as strong a dissipation, with the shown low-rank adaptive HyperDeepONets further reducing the error of the full HyperDeepONet, leading to overall much improved solutions for the linear advection equation~\eqref{eq:LinearAdvection}.

\subsection{Burgers equation}

As a next example we consider the viscous Burgers equation
\begin{equation}\label{eq:BurgersEqn}
 u_t + uu_x -\nu u_{xx} = 0,
\end{equation}
where we choose the small $\nu=0.025/\pi$ as diffusion coefficient to allow for the formation of steep solution gradients. We solve the problem again over the spatio-temporal domain $[-\pi,\pi]\times[0,t_{\rm f}]$, with $t_{\rm f}=1$. As for the linear advection equation, the initial conditions are sampled from a truncated Fourier series, and periodic boundary conditions are being used. For this example we experiment with enforcing the initial condition as a soft constraint using $\lambda_{\mathsf I}=100$ in the physics-informed loss function~\eqref{eq:compositeLossFunctionIBVP}, rather than as a hard constraint, to assess the ability of the HyperDeepONets to successfully minimize a more complicated composite loss function. 

The standard DeepONet is designed to have both 4 trunk and branch network layers with 128 units per layer each. The HyperDeepONet uses 4 trunk layers with 20 units per layer and 5 branch layers with 66 units per layer. The architectures for this problem are again inspired by those chosen in~\cite{lee23a}. 

\begin{table}[!ht]
\centering
\caption{Numerical results for Burgers equation~\eqref{eq:BurgersEqn} over the testing dataset.}\label{tab:BurgersEqn}
\begin{tabular}{llrl}
\toprule
Model & LoRA Rank & Params & $l_2$-error \\
\midrule
DeepONet & N/A & 116096 & $0.3161 \pm 0.1051$ \\
HyperDeepONet & N/A & 117389 & $0.1130 \pm 0.0044$ \\
LoRA-HyperDeepONet & 8 & 38979 & $0.1059 \pm 0.0113$ \\
LoRA-HyperDeepONet & 16 & 50395 & $0.1069 \pm 0.0072$ \\
LoRA-HyperDeepONet & 32 & 73227 & $\textbf{0.1053} \pm \textbf{0.0093}$ \\
\bottomrule
\end{tabular}
\end{table}

The results for the Burgers equation investigation are collected in Table~\ref{tab:BurgersEqn}. All of the HyperDeepOnet vastly improve upon the standard DeepONet, which is in line with the analogous results found in~\cite{lee23a} for the data-driven case. This is to be expected, as the HyperDeepONets were specifically designed for complex solutions, such as those exhibited by Burgers equation in the low viscosity regime, which leads to the formation of shocks. As in the previous examples, also for the Burgers equation suitable low-rank adaptive HyperDeepONets can outperform the full HyperDeepONet, albeit by a rather small margin. In this particular example we found the rank 32 LoRA-HyperDeepONet to perform the best, closesly followed by the rank 8 and 16 LoRa-HyperDeepoNets, which achieve comparable performance at less than half of the number of trainable parameters. In general, with just 30\%--50\% of the parameters, the LoRA-HyperDeepONets can slightly outperform the full HyperDeepOnet, making the low-rank adaptive versions again much more computationally efficient. 

\begin{figure}[!ht]
    \centering
    \includegraphics[width=\linewidth]{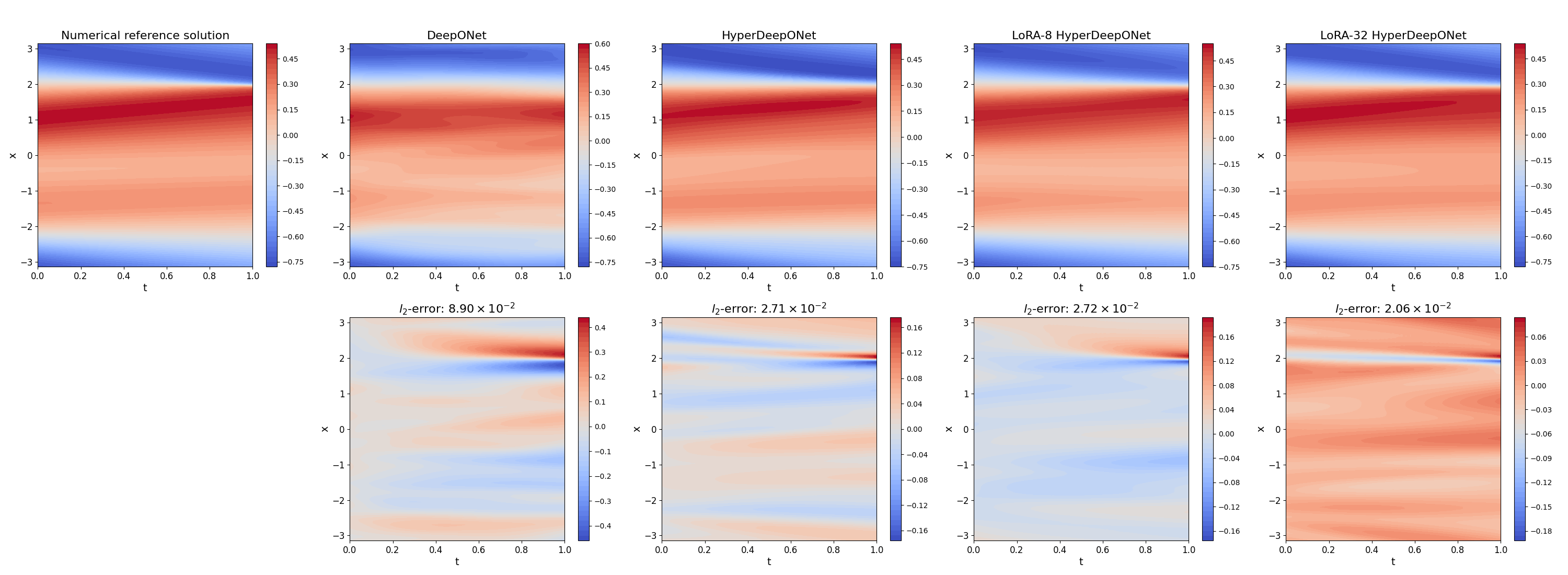}
    \caption{A case study for the viscous Burgers equation~\eqref{eq:BurgersEqn} for all the trained DeepONets evaluated over a single time-step starting from a randomly sampled initial condition $u_0(x)$ obtained from a truncated Fourier series expansion. \textit{Top row:} Numerical reference solution and DeepONet solutions. \textit{Bottom row:} Point-wise differences between the reference solution and the DeepONet solutions.}
    \label{fig:BurgersEqn}
\end{figure}

Figure~\ref{fig:BurgersEqn} illustrates the performance of the trained DeepONets for a randomly sampled initial condition $u_0(x)$. This figure clearly illustrates that the standard DeepONet is struggling to resolve the shock formation, and smooths it out instead. The HyperDeepONets are much more effective at resolving these shocks, again leading to overall lower errors for this benchmark. 

\subsection{Shallow-water equations}

As a last example we consider the one-dimensional shallow-water equations. This is a hyperbolic system of two first order equations,
\begin{align}\label{eq:ShallowWater}
 u_t + uu_x + gh_x =0,\quad h_t + (uH)_x = 0
\end{align}
where $H=h_0+h$ is the total water height, $h_0$ is a constant reference level, here chosen as $h_0=10$, and $h=h(t,x)$ is the displacement height from that reference level. The horizontal velocity is $u=u(t,x)$ and $g=9.81$ is the gravitational constant. This system of equations is solved over the spatio-temporal domain $[-\pi,\pi]\times[0,t_{\rm f}]$ with $t_{\rm f}=0.1$. Again, we use periodic boundary conditions, which are enforced as hard constraints, and so are the initial conditions, which for $h_0(x)$ are drawn from a class of Gaussian with amplitude $A$, shift $x_0$ and width $\sigma$. We sample $A$, $x_0$ and $\sigma$ uniformly randomly from the intervals $[0.5,1.5]$, $[-\pi/2,\pi/2]$ and $[0.5, 1.0]$, respectively. We set $u_0(x)=0$ for all initial conditions.

Our standard DeepONet consists of a branch network with 4 layers and 127 units per layer and a trunk network with 4 layers and 40 units per layer. The HyperDeepOnet utilizes a branch network of 4 layers and 20 units per layer and also a trunk network of 4 layers and 40 units per layer. This choice yields once more a similar parameter count of the DeepONet and full HyperDeepONet neural networks.

\begin{table}[!ht]
\centering
\caption{Numerical results for the shallow-water equations~\eqref{eq:ShallowWater} over the test dataset.}\label{tab:ShallowWaterEqns}
\begin{tabular}{llrl}
\toprule
Model & LoRA Rank & Params & $l_2$-error \\
\midrule
DeepONet & N/A & 114750 & $0.5918 \pm 0.0285$ \\
HyperDeepONet & N/A & 114802 & $0.0126 \pm 0.0023$ \\
LoRA-HyperDeepONet & 4 & 32290 & $0.0119 \pm 0.0034$ \\
LoRA-HyperDeepONet & 8 & 53018 & $0.0116 \pm 0.0028$ \\
LoRA-HyperDeepONet & 16 & 94474 & $\textbf{0.0110} \pm \textbf{0.0022}$ \\
\bottomrule
\end{tabular}
\end{table}

The results from the evaluation of the trained models for the shallow-water equations are presented in Table~\ref{tab:ShallowWaterEqns}. As for the other two partial differential equations examples also for the shallow-water equations, the HyperDeepOnets can vastly outperform the standard DeepONet. The overall lowest error is once more achieved by the low-rank adaptive versions of the HyperDeepONet, here corresponding to the rank $r=16$, although the low-rank versions corresponding to ranks $r=4$ and $r=8$ also outperform the full HyperNetwork, again at a lower total parameter count.

\begin{figure}[!ht]
    \centering
    \includegraphics[width=\linewidth]{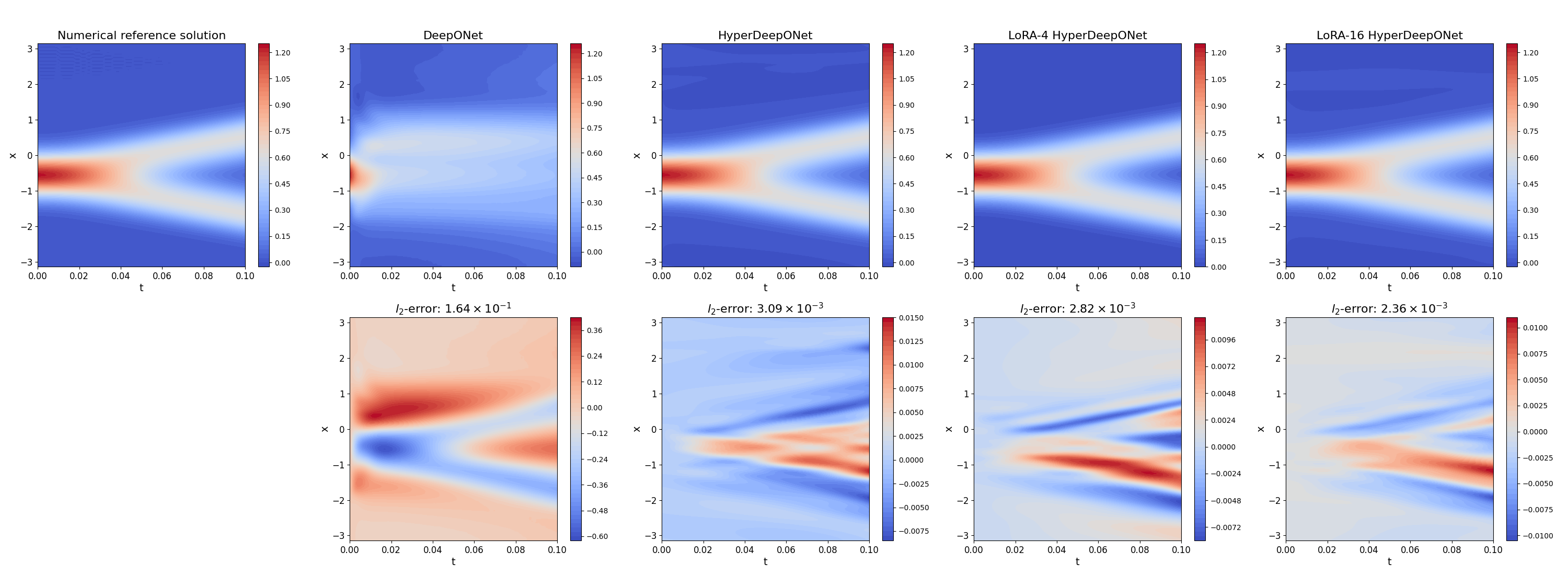}
    \caption{A case study for the one-dimensional shallow-water equations~\eqref{eq:ShallowWater} for all the trained DeepONets evaluated over a single time-step starting from a randomly sampled initial condition with $u_0(x)=0$ and $h_0(x)$ being sampled from a Gaussian with random amplitude, shift and width. \textit{Top row:} Numerical reference solution and DeepONet solutions. \textit{Bottom row:} Point-wise differences between the reference solution and the DeepONet solutions.}
    \label{fig:SWE}
\end{figure}

In Figure~\ref{fig:SWE} we present a case study for the shallow-water equations evaluating the trained DeepONets. The standard DeepONet visually struggles to accurately capture the evolution of the outward moving wave fronts, while the HyperDeepONets correctly capture this behaviour. The overall lowest errors are again achieved by the low-rank adaptive HyperDeepONets, similar as in the previous examples.

\section{Conclusion}\label{sec:Conclusion}

We have introduced a parameter-efficient approximation to the HyperDeepONets introduced in~\cite{lee23a}, by defining a low-rank adaptivity approximation for the weight matrix connecting the last hidden layer and the output layer of the branch network of these HyperDeepONets. As a further novel contribution, while the paper~\cite{lee23a} assessed the performance of HyperDeepONets in a purely data-driven setting, we have considered the case of physics-informed machine learning here. That is, no training data had to be generated offline, and we can directly learn the solution operator of our systems of differential equations solely from a predefined class of initial--boundary conditions using a physics-informed loss function.

It is important to acknowledge a limitation in the direct comparison between standard DeepONets and HyperDeepONets. The vast design space allows for numerous ways to allocate a fixed parameter budget between the branch and trunk networks, and we could not exhaustively explore all possible configurations. However, the comparison between the full HyperDeepONet and our proposed LoRA-HyperDeepONets is direct, as they share the same underlying architecture. The consistent improvements observed with the LoRA variants, therefore, are not an artifact of architectural choices but a direct result of the low-rank adaptation. In particular, extensive testing within this physics-informed machine learning setting have revealed that the proposed low-rank adaptive HyperDeepONets can match or outperform full HyperDeepONets at a fraction of the neural network parameters required. In particular, we experimented with different architectures such as shallow and wide as well as deep and narrow for both trunk and branch networks, soft- and hard-constrained initial conditions, both ordinary and partial differential equations, and one-step and iterative $n$-step inference. Across all these experiments low-rank adaptive HyperDeepONets have emerged as a robust alternative to both vanilla DeepONets and full HyperDeepONets, for equations with both simple (such as periodic or advected) and more complicated (such as breaking waves) solutions. We find these results surprising as the low-rank adaptive HyperDeepONets can achieve competitive performance to standard DeepONets and full HyperDeepONets at a fraction of the required trainable parameters. This implies that constraining the output matrix of the HyperDeepONet's branch network to not have full rank acts as a regularization mechanism that simplifies the physics-informed loss surface, thus facilitating convergence to better solutions to the required optimization problem. 

Our work also highlights several promising avenues for future research. While the proposed low-rank adaptive method proved effective, the rank $r$ itself remains a fixed hyperparameter that requires tuning. As evidenced by the examples presented in this paper, the optimal rank $r$ is different for each system of differential equations, and due to computational limitations, not all values for $r$ could be tested exhaustively in this study. A significant advancement would be to explore methods for adapting the rank dynamically during the training process, thereby allowing the LoRA-HyperDeepONet training algorithm to automatically allocate capacity based on the complexity of the solution operator. Furthermore, while the tested examples included challenging nonlinear dynamics and the formation of steep gradients, future work should investigate the performance of LoRA-HyperDeepONets on more complex, multi-scale phenomena such as turbulent flows. Extending this parameter-efficient framework to such problems and to systems in higher spatial dimensions would be a valuable next step that could further assess the benefits of this approach.

\begin{ack}
This research was carried out, in part, through funding from the Canada Research Chairs program and the NSERC Discovery Grant program.
\end{ack}

{\footnotesize\setlength{\itemsep}{0ex}

}

\end{document}